\documentclass[10pt]{article} 

\usepackage[preprint]{sty}

\usepackage{hyperref}
\usepackage{url}

\usepackage{amsmath}
\usepackage{amssymb}

\usepackage{wrapfig}
\usepackage{graphicx}
\usepackage{booktabs}
\usepackage{multirow}
\usepackage{array}
\usepackage[table]{xcolor}
\definecolor{mygray}{gray}{0.4}
\definecolor{lightbluecell}{RGB}{230,240,255}

\newcommand{\bluecell}{\cellcolor{lightbluecell}}

\newcommand{\addrow}{\makebox[1.2em][c]{+}}
\newcommand{\subrow}{\makebox[1.2em][c]{$-$}}

\makeatletter
\newcommand\notsotiny{\@setfontsize\notsotiny\@vipt\@viipt}
\makeatother

\usepackage{subcaption}

\title{Rethinking Classifier-Free Guidance for TI2I Flow Matching}

\newcommand{\imperial}{\textsuperscript{\rm $\star$}}
\newcommand{\ms}{\textsuperscript{\rm \textdaggerdbl}}

\author{%
\makebox[\textwidth][c]{%
\parbox{\textwidth}{\centering
Andreas Floros\textsuperscript{\rm $\star$, \textdaggerdbl}\quad
Elena Koritska\ms\quad
Teofilo de Campos\ms\\
Christoph Vogel\ms\quad
Chu-Tak Li\ms\quad
Anuj Sharma\ms\quad
Alexandros Neophytou\ms\\[0.6em]
\textmd{\imperial Imperial College London \qquad \ms Microsoft}\\[0.3em]
\textmd{\texttt{andreas.floros18@imperial.ac.uk}}
}%
}%
}

\def\openreview{\url{https://openreview.net/forum?id=XXXX}} 

\begin{document}

\maketitle

\begin{abstract}
We study Text-and-Image-to-Image (TI2I) flow matching methods for controllable image editing. In a broader generative modeling context, Classifier-Free Guidance (CFG) is established as a key component responsible for high quality outputs. However, we argue that its role in TI2I modeling is poorly understood. Specifically, we make the surprising observation that the CFG mechanism is implicitly harmful for editing in the noisy phase of flow modeling, hindering instruction following and controllability. We propose a simple correction in the early flow steps, enabling a richer alignment-preservation tradeoff at no extra cost. The effectiveness of our approach is demonstrated on both pixel-space and latent TI2I flows.
\end{abstract}

\begin{figure}[h]
    \centering
    \includegraphics[width=\linewidth]{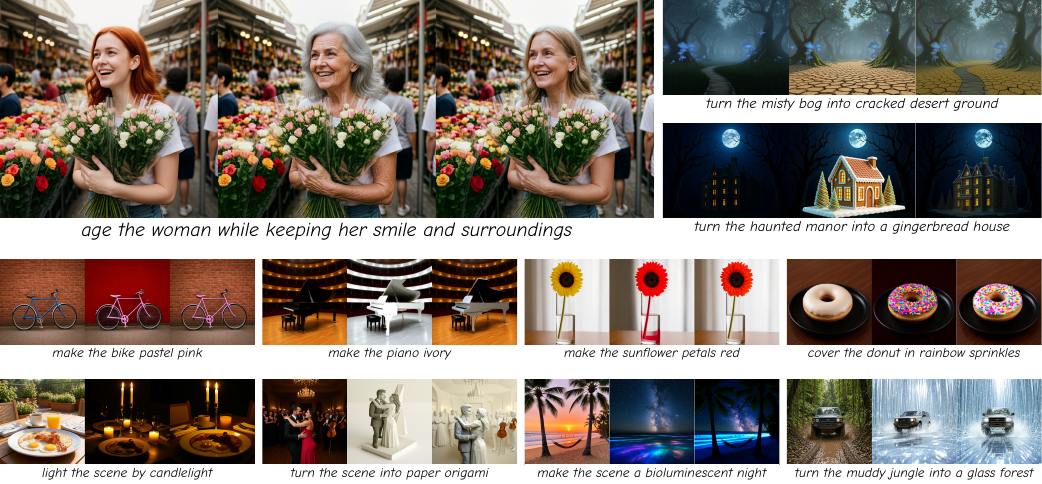}
    \caption{We propose a simple modification in TI2I flows that substantially improves source preservation while maintaining text alignment. In the examples above, source images are on the left, baseline edits in the middle (we evaluate Qwen, FLUX.2 and PixelGen models) and editing with our correction on the right.}
    \label{fig:teaser}
\end{figure}

\section{Introduction}
\label{sec:introduction}

Text-and-Image-to-Image (TI2I) editing is the task of modifying a source image according to a text prompt while retaining the source wherever no change is intended. The source image provides the concrete visual instance, including scene, object identity, composition, appearance and incidental details. The prompt instead serves as an instruction specifying how that instance should change. Crucially, a successful edit must apply the requested changes without reopening the rest of the image for synthesis.

TI2I editing methods studied in the literature typically adapt pretrained Text-to-Image (T2I) models. These methods are often training-free, so their design space concentrates on inference-time mechanisms \citep{Kulikov_2025_ICCV, yang2026directeditsteplevelaccurateinversion}. This adaptation paradigm has produced a broad and useful set of techniques but it also inherits limitations from models trained primarily for T2I generation. In parallel, a newer line of systems trains text-and-image-conditioned flows directly for editing \citep{wu2025qwenimagetechnicalreport, flux-2-2025}. These models are increasingly important yet their properties remain comparatively underexplored.

In this paper, we study this setting of bespoke TI2I flow matching, rather than editing through adaptation of a T2I generator. Specifically, we focus on the nuance of Classifier-Free Guidance (CFG), which is a central mechanism of modern generative flows \citep{ho2021classifierfree}. In T2I generation, increasing the CFG weight, $\omega$, is commonly understood as improving text alignment at the cost of sample diversity and, at large values, visual naturalness \citep{zheng2024characteristic, karras2024guiding}. In editing, however, the relevant tradeoff is alignment versus preservation with respect to the source. In that case, we find that CFG produces a poorly shaped guidance tradeoff. That is, small $\omega$ weakens instruction following but may not reliably reduce distortion, while large $\omega$ is often essential for alignment but tends to amplify unnecessary changes.

We identify the source of this inefficiency and attribute it to the well-established and widely adopted recipe of \citet{ho2021classifierfree}. As demonstrated in Figure \ref{fig:teaser}, we propose a simple inference-time correction that largely mitigates the issue. Our intervention requires no additional evaluations, leaves the trained model unchanged and recovers a richer set of operating points as $\omega$ varies. We show this in two settings. For a controlled analysis, we use a PixelGen \citep{ma2026pixelgenimprovingpixeldiffusion} finetune for pixel-space editing. We then demonstrate our approach in more general editing, benchmarking via Qwen-Image-Edit \citep{wu2025qwenimagetechnicalreport} and FLUX.2 \citep{flux-2-2025}, representative off-the-shelf latent flows tailored for the task.

\section{Background}

We give a brief overview of literature on modern generative flows, techniques for TI2I modeling and CFG.

\subsection{Flow Matching}
\label{subsec:flowmatching}

Flows perform generative modeling through Ordinary Differential Equations (ODEs) \citep{chen2018cnf}. Letting $p$ represent the target distribution on $\mathbb{R}^D$, they correspond to the following transport dynamics:
\begin{equation}
    \label{eq:ode}
    \text{d}\boldsymbol{x}_t=\boldsymbol{v}_{\boldsymbol{\theta}}(\boldsymbol{x}_t,t)\text{d}t, \quad t\in[0,1], \quad \boldsymbol{x}_1\sim p,
\end{equation}
where $\boldsymbol{v}_{\boldsymbol{\theta}}:\mathbb{R}^D\times\mathbb{R}\to\mathbb{R}^D$ denotes a velocity field parameterized by $\boldsymbol{\theta}$. In principle, there are several transport paths that satisfy Equation \ref{eq:ode}. In practice, \citet{lipman2023flow} propose to model a linear interpolation between a tractable standard Gaussian distribution and the desired $p$. The ODE then defines marginal distributions, $q_t$, subject to the constraints. The corresponding flow matching objective is as follows:
\begin{equation}
    \min_{\boldsymbol{\theta}}\mathbb{E}_{t,\boldsymbol{x}_1\sim p, \boldsymbol{x}_0\sim \mathcal{N}(\boldsymbol{0},\boldsymbol{I})}\left\{\left\|\boldsymbol{v}_{\boldsymbol{\theta}}\left[\psi_t(\boldsymbol{x}_0;\boldsymbol{x}_1),t\right]-\frac{\text{d}}{\text{d}t}\psi_t(\boldsymbol{x}_0;\boldsymbol{x}_1)\right\|^2_2\right\}, \quad \psi_t(\boldsymbol{x}_0;\boldsymbol{x}_1):=(1-t)\boldsymbol{x}_0+t\boldsymbol{x}_1,
\end{equation}
where $\psi_t$ denotes the conditional flow and its time derivative yields the associated conditional velocity. Note that this formulation is closely related to score-based generative models \citep{song2021scorebased}. That is, letting $\text{Id}$ denote the identity map, scores are expressed in terms of velocity as $\nabla_{\boldsymbol{x}}\log q_t=(1-t)^{-1}[t\boldsymbol{v}_{\boldsymbol{\theta}}(\cdot,t)-\text{Id}]$.

\subsection{Text-and-Image-to-Image Modeling}

TI2I models generate edit candidates conditioned jointly on a source image, $\boldsymbol{c}^{\text{img}}$, and a text prompt or instruction $\boldsymbol{c}^{\text{txt}}$. That is, the target distribution here is of the form $p(\cdot\mid\boldsymbol{c}^{\text{img}},\boldsymbol{c}^{\text{txt}})$. Unlike T2I generation, which has weaker conditioning, sampling from $p(\cdot\mid\boldsymbol{c}^{\text{txt}})$, TI2I editing requires balancing two competing guiding signals and goals. Namely, adherence to the requested edit and preservation of source content.

Most editing methods in the literature do not train TI2I editing models directly. Instead, they adapt pretrained T2I models at inference time, effectively exploiting their strong generative priors while avoiding additional optimization costs \citep{meng2022sdedit, rout2025semantic, Kulikov_2025_ICCV, yang2026directeditsteplevelaccurateinversion}.

Alternatively, it is also possible to train a generator directly for image editing. Early diffusion-based examples include \citet{brooks2023instructpix2pix}, which learns to follow natural-language editing instructions from paired editing data. More recently, large-scale multimodal flow models have been trained natively with both image and text conditioning, enabling direct modeling of the TI2I distribution \citep{wu2025qwenimagetechnicalreport, flux-2-2025}. Notably, in these systems, the source image enters the model as a first-class conditioning signal. In particular, for flow matching models, conditioning is typically incorporated into the velocity field, $\boldsymbol{v}_{\boldsymbol{\theta}}(\boldsymbol{x}_t,t,\boldsymbol{c}^{\text{img}},\boldsymbol{c}^{\text{txt}})$, so that the sampling trajectory is jointly determined by the current state, source image and textual instruction.

\subsection{Classifier-Free Guidance}

\citet{ho2021classifierfree} introduced CFG as a principled way to trade mode coverage for alignment in conditional models. Intuitively, CFG aims to sample from tilted distributions, $q_{t, \omega}(\cdot\mid\boldsymbol{c})\propto q^{1-\omega}_t(\cdot\mid\varnothing)q^{\omega}_t(\cdot\mid\boldsymbol{c})$, where the guidance weight, $\omega$, controls the tradeoff and $\varnothing$ denotes an empty condition. Note, this corresponds to a linear interpolation of score functions. That is, taking logarithms and differentiating, we write:
\begin{equation}
    \nabla_{\boldsymbol{x}}\log q_{t, \omega}(\cdot\mid\boldsymbol{c})=(1-\omega)\nabla_{\boldsymbol{x}}\log q_t(\cdot\mid\varnothing)+\omega\nabla_{\boldsymbol{x}}\log q_t(\cdot\mid\boldsymbol{c}).
\end{equation}
Therefore, given the equivalence of scores with velocities, the implementation requires training both an unconditional and a conditional model. In practice, this is done by dropping conditioning information during training for a fraction of the iterations, effectively unifying the two flows under a single parameterization. Despite the theoretical elegance of CFG, subtle technical details make it a non-trivial mechanism, with more nuanced dynamics and interpretation than the original motivation suggests \citep{zheng2024characteristic, karras2024guiding}. Moreover, its design in the presence of multimodal conditioning remains relatively unexplored.

\section{Methodology}

Before proceeding with analysis, we summarize our experimental setup. For completeness, we consider both pixel-space and latent-space TI2I models. In particular, pixel-space modeling enables a controlled and more interpretable setup. To this end, we produce a TI2I finetune of PixelGen \citep{ma2026pixelgenimprovingpixeldiffusion} at $512\times512$ resolution. For latent-space modeling, we experiment with Qwen-Image-Edit \citep{wu2025qwenimagetechnicalreport} and the FLUX.2 family \citep{flux-2-2025} at $256\times256$, $512\times512$ and $1024\times1024$ image resolutions.

Throughout the paper, we use synthetic datasets, created by existing foundation models. After quality control, each contains approximately 200 entries. Specifically, we first generate text triplets of source image descriptions, edit instructions and post-edit descriptions via Claude Opus 4.7. Source descriptions are then used to generate source images via FLUX.2. Additionally, for local editing with PixelGen, we constrain the text triplets to describe scenarios where changes are expected to be strictly confined within specific regions in the images. In that case, we use the Segment Anything Model (SAM) of \citet{carion2026sam} to automatically obtain corresponding segmentation masks. For latent-space editing with Qwen and FLUX.2, which we found to be significantly more capable, we allow general and challenging triplets to effectively test limitations.

Further details on infrastructure, data, model configurations and hyperparameters are given in Appendix \ref{appendix:setup}.

\subsection{Understanding Guidance \& Its Design Space in TI2I Flows}
\label{subsec:profiles}

Our goal is understanding when editing failures emerge and how to interpret them, with the expectation that this will lead to actionable insights. Concretely, here we investigate TI2I flow modeling under CFG. Following the equivalence with score matching, this corresponds to the following ODE dynamics:
\begin{equation}
    \label{eq:cfgode}\text{d}\boldsymbol{x}_t=\boldsymbol{v}_{\boldsymbol{\theta},\omega}(\boldsymbol{x}_t,t,\boldsymbol{c})\text{d}t, \quad \boldsymbol{v}_{\boldsymbol{\theta},\omega}(\cdot,t, \boldsymbol{c}):=\boldsymbol{v}_{\boldsymbol{\theta}}(\cdot,t,\varnothing)+\omega\Delta\boldsymbol{v}_{\boldsymbol{\theta}}(\cdot,t,\boldsymbol{c}), \quad \boldsymbol{x}_0\sim \mathcal{N}(\boldsymbol{0},\boldsymbol{I}),
\end{equation}
where we abbreviate joint conditioning as $\boldsymbol{c}=\boldsymbol{c}^{\text{img}},\boldsymbol{c}^{\text{txt}}$ and $\Delta\boldsymbol{v}_{\boldsymbol{\theta}}(\cdot,t,\boldsymbol{c}):=\boldsymbol{v}_{\boldsymbol{\theta}}(\cdot,t,\boldsymbol{c}^{\text{txt}})-\boldsymbol{v}_{\boldsymbol{\theta}}(\cdot,t,\varnothing)$ is the differential velocity. Importantly, following the documentation of modern TI2I models \citep{wu2025qwenimagetechnicalreport, flux-2-2025}, CFG is only applied with respect to text conditioning. That is, all terms appearing in $\boldsymbol{v}_{\boldsymbol{\theta},\omega}(\cdot,t, \boldsymbol{c})$ and $\Delta\boldsymbol{v}_{\boldsymbol{\theta}}(\cdot,t,\boldsymbol{c})$, where we have slightly abused notation for compactness, should be understood as being implicitly conditioned on $\boldsymbol{c}^{\text{img}}$. In that case, $\varnothing$ represents an empty text prompt and text-unconditional modeling, that remains conditioned on $\boldsymbol{c}^{\text{img}}$. We adopt these conventions for the remainder of the paper.

\begin{figure}[t]
    \centering
    \includegraphics[width=\linewidth]{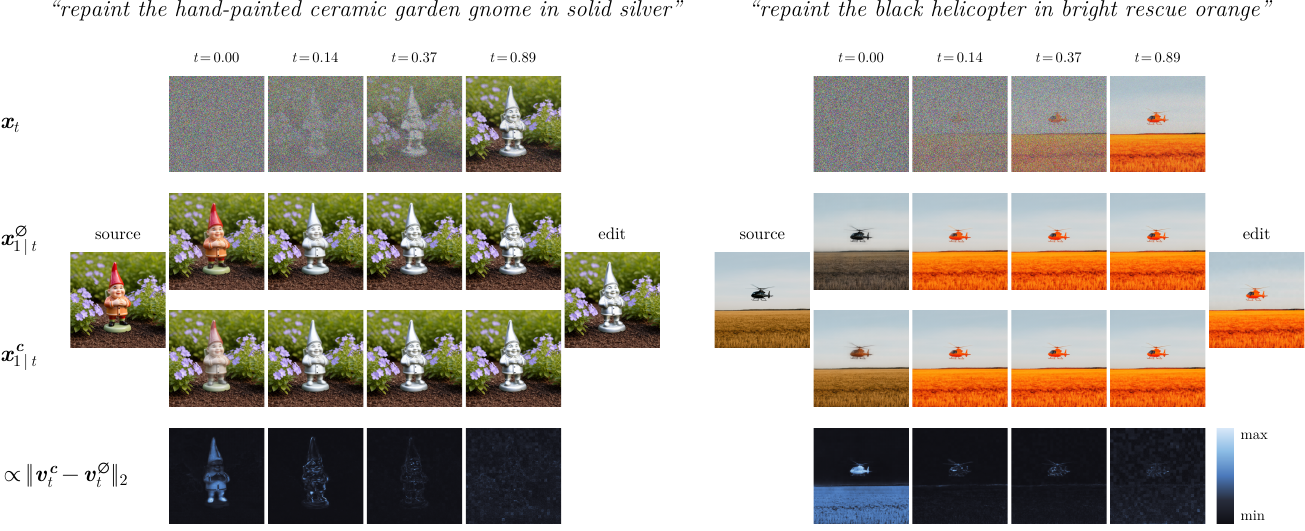}
    \caption{Visualization of two representative TI2I modeling scenarios with our PixelGen finetune. The first rows show the iterates $\boldsymbol{x}_t$. The second and third rows show the unconditional and conditional predictions of the network for the output edit image given $\boldsymbol{x}_t$, which can be derived given knowledge of the corresponding velocities. The last rows show the evolution of the differential velocity, equivalent to the difference of the second and third rows. Specifically, we visualize its magnitude as a heatmap that is normalized for each $t$. In both cases, the results of the last row concentrate on the edit regions near $t=0$ and become less localized and noisy as $t\to1$. Moreover, the failure in the right example is evident immediately at $t=0$ and we can intuitively attribute it to the unconditional prediction of the second row, where the field has been repainted.}
    \label{fig:trajectories}
\end{figure}

\begin{wrapfigure}{r}{0.4\textwidth}
    \centering
    \includegraphics[width=0.4\textwidth]{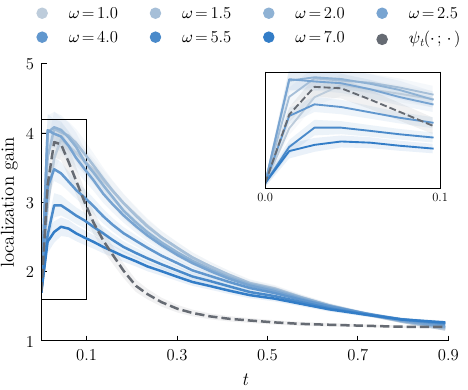}
    \caption{Average localization gain ($\pm$ the standard error) for our PixelGen finetune. We use SAM 3 \citep{carion2026sam} for the oracle masks. Results labeled as $\psi_t(\cdot;\cdot)$ correspond to using the analytical flow model, previously discussed in Section \ref{subsec:flowmatching}.}
   \label{fig:localization_gain}
\end{wrapfigure}

Specifically for this section, we focus on pixel-space and strictly local editing with PixelGen. Our motivation for this is two-fold. First, since transport dynamics unfold in data space here, we are able to gain visual intuition for how flow modeling works by directly inspecting intermediate results. Equipped with this knowledge, we may then test our findings in more general environments, not amenable to the same analysis. In fact, this enables our controlled local editing analysis. Importantly, our second argument for studying pixel-space models is that we avoid confounders that emerge from autoencoding and latent-space design, ensuring that our analysis is grounded in dynamics arising solely due to flow modeling.

Having motivated our setting, we set out to study the fundamental quantity of CFG in Equation \ref{eq:cfgode}, namely, the differential velocity $\Delta\boldsymbol{v}_{\boldsymbol{\theta}}(\cdot,t,\boldsymbol{c})$. As a first step, we visualize its evolution along with intermediate results over flow modeling in Figure \ref{fig:trajectories}. The examples we have selected here are representative of a general trend. That is, we observe that the differential velocity is localized in the early steps of flow modeling, becoming less local and noisy as $t\to1$. In particular, we are able to quantify this phenomenon via the localization gain, defined as follows:
\begin{equation}
    \text{LG}(\cdot,t,\boldsymbol{c}):=
\mathbb{E}_{u\sim\mathcal{U}(\Omega_{M})}\!\bigl\|[\Delta\boldsymbol{v}_{\boldsymbol{\theta}}(\cdot,t,\boldsymbol{c})]_u\bigr\|_2
\;\Big/\;
\mathbb{E}_{u\sim\mathcal{U}(\Omega)}\!\bigl\|[\Delta\boldsymbol{v}_{\boldsymbol{\theta}}(\cdot,t,\boldsymbol{c})]_u\bigr\|_2,
\end{equation}
where $\Omega$ represents the image (pixel) domain, the set $\Omega_{M}\subseteq\Omega$ is determined via an oracle binary mask, $M$, with no support outside the intended edit region and
$[\,\cdot\,]_u$ is evaluation at pixel $u$. In words, LG compares the mean per-pixel magnitude of $\Delta\boldsymbol{v}_{\boldsymbol{\theta}}(\cdot,t,\boldsymbol{c})$ inside the edit region against its mean over the
complete image.

If $\text{LG}\approx 1$ then the differential velocity is spatially uniform whereas $\mathrm{LG}\gtrless1$ implies it concentrates inside or outside of $\Omega_{M}$ respectively. To investigate, in Figure \ref{fig:localization_gain} we report this quantity averaged across our dataset as a function of $t$. The curves confirm quantitatively what
Figure \ref{fig:trajectories} suggests. Namely, LG generally decreases as $t\to 1$, with $\Delta\boldsymbol{v}_{\boldsymbol{\theta}}(\cdot,t,\boldsymbol{c})$ becoming increasingly uniform. Curiously, however, we see that localization consistently rises from $t=0$ before peaking early. To probe whether this profile reflects the trained model or is prescribed by our particular sampling and discretization procedure, we additionally evaluate
$\mathrm{LG}$ along the analytical flow, $\psi_t(\cdot;\cdot)$, from
Section \ref{subsec:flowmatching}, where $\boldsymbol{x}_t$ is drawn
directly from the linear interpolant rather than rolled out by the
reverse ODE. In that case, we observe a similar trend, i.e., the temporal profile of LG is inherent to the model, not an artifact of a particular trajectory taken by sampling.

We conclude this section by emphasizing the key findings of the above experiments. First, our locality analysis suggests that edit-relevant signal is concentrated in the noisy phase. That is, the spatial structure of $\Delta\boldsymbol{v}_{\boldsymbol{\theta}}(\cdot,t,\boldsymbol{c})$ effectively encodes where the edit will land before any detail has been resolved in $\boldsymbol{x}_t$. Second, as Figure \ref{fig:trajectories} illustrates anecdotally, intermediate predictions in this same phase are already predictive of the final outcome, i.e., we expect a successful edit and an over-aggressive one to diverge early. Taken together, this identifies the noisy phase as the regime in which TI2I editing is effectively decided, and in which any corrective intervention is likely to be both necessary and sufficient. We turn to such an intervention next.

\subsection{The Unconditional Branch Pathology and a Simple Remedy}

One interpretation of non-monotonic localization profiles observed in Figure \ref{fig:localization_gain} is that they reflect a modeling defect. Specifically, edit failures, as shown on the right in Figure \ref{fig:trajectories}, may substantially degrade the average LG in early stages of flow modeling by placing support outside of $\Omega_{M}$. Then, as $t$ increases, reparation of the profiles intuitively requires dispersing this excess mass, causing LG to initially increase before collapsing. Given this reasoning and the analysis of Section \ref{subsec:profiles}, suggesting that the observed trend is independent from sampling dynamics, we are therefore motivated to investigate model-intrinsic failure cases near $t=0$.
\begin{figure}[t]
\begin{minipage}{0.5\textwidth}
    \centering

\includegraphics[width=0.8\textwidth]{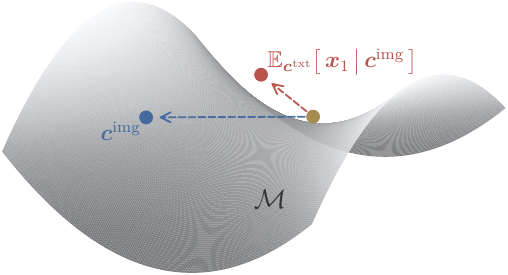}
\caption{Consider a data manifold, $\mathcal{M}\subseteq\mathbb{R}^D$, and an arbitrary point in ambient space. The theoretical text-unconditional velocity, in blue, pushes points on $\mathcal{M}$, toward source $\boldsymbol{c}^{\text{img}}$. The CFG-trained velocity, in red, steers toward the mean edit, which need not $\in\mathcal{M}$.}
\label{fig:manifold}
\end{minipage}
\hfill
\begin{minipage}{0.45\textwidth}
    \centering

\includegraphics[width=\textwidth]{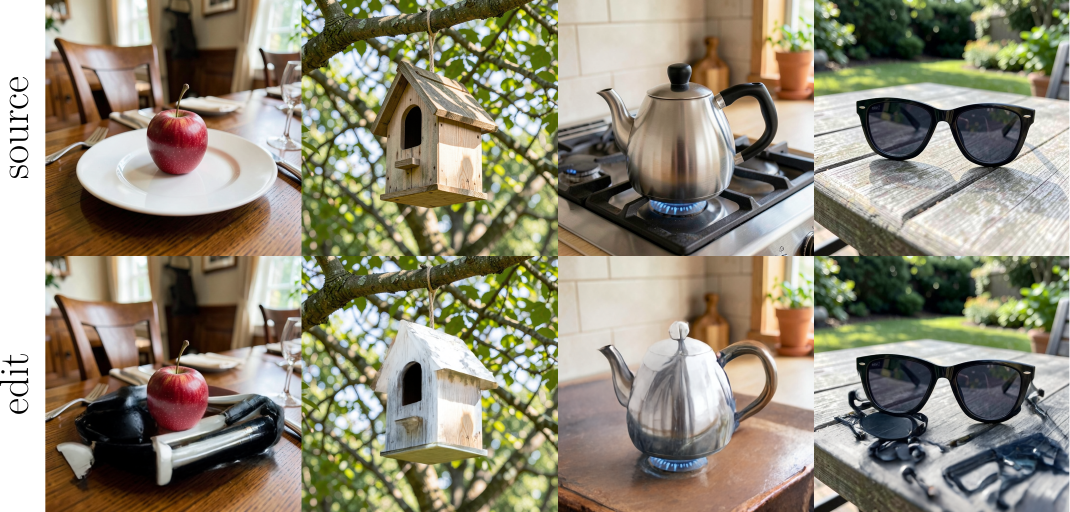}
\caption{Results with our TI2I PixelGen finetune for text-unconditional editing, equivalent to setting the CFG weight $\omega=0$. The top row shows source, input images. The outputs, shown on the bottom row, are distorted and we do not have faithful reconstruction of the inputs.}
\label{fig:recon}
\end{minipage}
\end{figure}

Our insight is that, while TI2I models are trained for editing, they are only implicitly taught to preserve the source's structure. In particular, we can formalize this phenomenon by examining the text-unconditional velocity, $\boldsymbol{v}_{\boldsymbol{\theta}}(\cdot,t,\varnothing)$, which, given no text instruction, we intuitively interpret as a no-edit flow toward the source. Concretely, for $t\approx0$, say $t<\tau$ for some $\tau>0$, we expect this to be well-approximated by $(\boldsymbol{c^{\text{img}}}-\text{Id})/(1-t)$, since noisy inputs in this regime carry little information. However, as illustrated in Figure \ref{fig:manifold}, this no-edit constraint is not necessarily captured during training. Instead, CFG trains the text-unconditional branch to approximate the same target as the conditional branch, effectively forcing an edit even in the absence of a text instruction. In that case, it follows that the unconditional branch steers data toward the mean edit $\mathbb{E}_{\boldsymbol{c}^{\text{txt}}}[\boldsymbol{x}_1\mid\boldsymbol{c}^{\text{img}}]$, which need not reside on the data manifold. As demonstrated in Figure \ref{fig:recon}, where we perform TI2I modeling with an empty text prompt, the results are therefore not guaranteed to be close to the source. That is, our analysis suggests that CFG implicitly hurts source preservation.

\begin{wrapfigure}{r}{0.45\textwidth}
    \centering
    \includegraphics[width=0.45\textwidth]{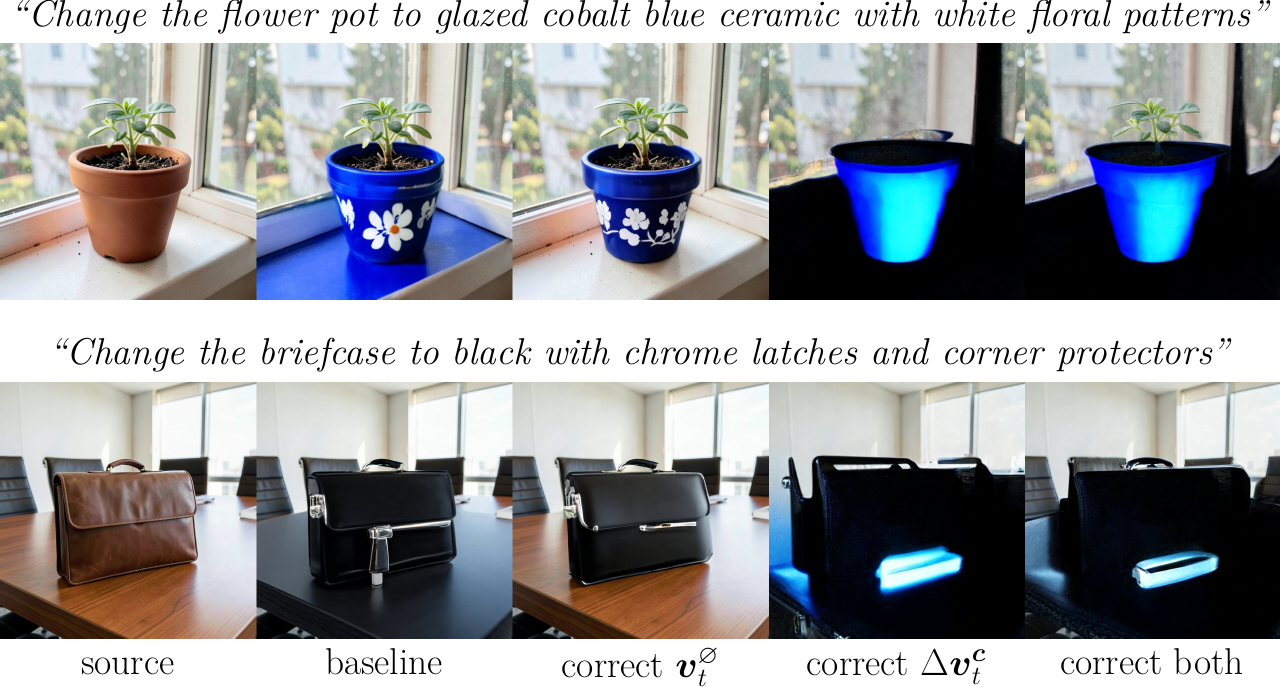}
    \caption{We ablate the proposed correction on our PixelGen finetune with simple edit tasks, setting $\tau=0.05$. Of tested configurations, only the one corresponding to Equation \ref{eq:earlyode} is successful.}
   \label{fig:ablation}
\end{wrapfigure}

To overcome this issue, we introduce a simple intervention, which we refer to as early correction in the remainder of the paper. Specifically, since we have an analytical approximation of the desired unconditional velocity for $t\approx0$ and given the findings of Section \ref{subsec:profiles}, where we empirically find that coarse structure is determined early in the flow modeling process, we propose to augment the dynamics of Equation \ref{eq:cfgode} as follows:
\begin{equation}
    \label{eq:earlyode}\text{d}\boldsymbol{x}_t=\boldsymbol{v}_{\boldsymbol{\theta},\omega,\tau}(\boldsymbol{x}_t,t,\boldsymbol{c})\text{d}t, \quad \boldsymbol{v}_{\boldsymbol{\theta},\omega,\tau}(\cdot,t, \boldsymbol{c}):=\boldsymbol{v}_{\boldsymbol{\theta},\tau}(\cdot,t,\varnothing)+\omega\Delta\boldsymbol{v}_{\boldsymbol{\theta}}(\cdot,t,\boldsymbol{c}), \quad \boldsymbol{x}_0\sim \mathcal{N}(\boldsymbol{0},\boldsymbol{I}).
\end{equation}
Here, $\boldsymbol{v}_{\boldsymbol{\theta},\tau}(\cdot,t,\varnothing)$ is the early-corrected unconditional branch that we define piecewise as:
\begin{equation}
    \label{eq:earlycorrected_unconditional}
    \boldsymbol{v}_{\boldsymbol{\theta},\tau}(\cdot, t, \varnothing) :=
    \begin{cases}
        \dfrac{\boldsymbol{c}^{\text{img}} - \text{Id}}{1 - t} & \text{if } t<\tau \\[8pt]
        \boldsymbol{v}_{\boldsymbol{\theta}}(\cdot, t, \varnothing) & \text{otherwise}.
    \end{cases}
\end{equation}
In practice, $\tau$ may be rounded to the nearest sampling step to satisfy a smoothness requirement \citep{kynkaanniemi2024applying}. Note that our proposed correction is valid for the isolated unconditional branch but not for the differential velocity $\Delta\boldsymbol{v}_{\boldsymbol{\theta}}(\cdot,t,\boldsymbol{c})$. We empirically verify this in Figure \ref{fig:ablation}, where the correction is successful only for our proposed configuration and otherwise results in severe distortion of the source image.

A practical explanation of this connects with \citet{karras2024guiding}, who find benefits of CFG can be interpreted as error cancellation. In that case, the unconditional branch, $\boldsymbol{v}_{\boldsymbol{\theta}}(\cdot,t,\varnothing)$, serves as a weak predictor of conditional velocity whereas the conditional branch, $\boldsymbol{v}_{\boldsymbol{\theta}}(\cdot,t,\boldsymbol{c}^{\text{txt}})$, is a stronger predictor, having smaller but directionally consistent errors with $\boldsymbol{v}_{\boldsymbol{\theta}}(\cdot,t,\varnothing)$, i.e., $\Delta\boldsymbol{v}_{\boldsymbol{\theta}}(\cdot,t,\boldsymbol{c})$ captures model-specific and meaningful errors that should be directionally preserved. In particular, full correction amounts to an implicit shift in the training distribution of the model, corresponding to a symmetrization of edit tasks such that $\mathbb{E}_{\boldsymbol{c}^{\text{txt}}}[\boldsymbol{x}_1\mid\boldsymbol{c}^{\text{img}}]=\boldsymbol{c}^{\text{img}}$, but this may not be well-approximated. Instead, our correction corresponds to a still valid tradeoff in terms of the symmetrized unconditional density and the preserved implicit CFG-induced classifier. That is, invoking the equivalence with score matching, roughly, we interpret this as sampling from a tilted distribution $q_{t, \omega, \tau}(\cdot\mid\boldsymbol{c})\propto q_{t,\tau}(\cdot\mid\varnothing)\left[\frac{q_t(\cdot\mid\boldsymbol{c})}{q_t(\cdot\mid\varnothing)}\right]^{\omega}$.

\subsection{Early Correction Unlocks a Richer Alignment-Preservation Tradeoff}

Having introduced early correction, we now turn to evaluation. As discussed in Section \ref{sec:introduction}, we are interested in quantifying the alignment-preservation tradeoff. We consider various proxies for each axis that are commonly adopted in the literature. Alignment is quantified by comparing the post-edit images with target post-edit descriptions via CLIP similarity \citep{pmlr-v139-radford21a}, PickScore \citep{kirstain2023pickapic} and HPSv3 \citep{Ma_2025_ICCV}. Separately, we use the Aesthetics predictor \citep{schuhmann2022laionaesthetics}, a no-reference, unconditional human preference proxy operating only on images. For preservation, we compare post-edit images with corresponding source images via PSNR, LPIPS \citep{lpips} and DINOv2 \citep{oquab2024dinov}.

\begin{wrapfigure}{r}{0.6\textwidth}
    \centering
    \includegraphics[width=0.6\textwidth]{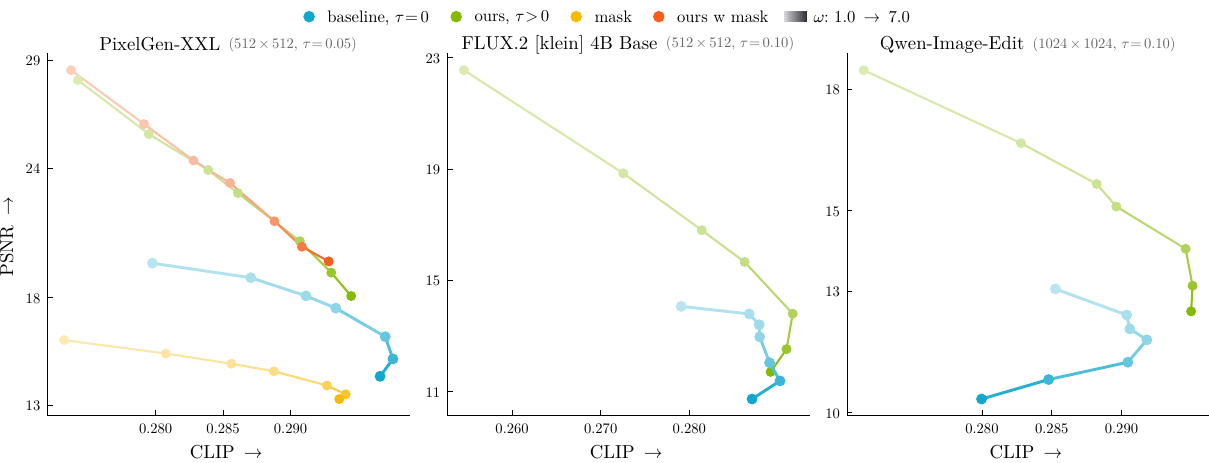}
    \caption{Alignment-preservation tradeoffs via PSNR and CLIP. We evaluate our PixelGen finetune, FLUX.2 and Qwen models. Curves form by varying $\omega$, with faint colors given to small values. Our method redefines the Pareto front, enabling robust editing.}
   \label{fig:tradeoff}
\end{wrapfigure}

In Table \ref{tab:results} and Figure \ref{fig:tradeoff} we report results with the above protocol. It is evident that early correction tends to improve performance across all models, with FLUX.2 especially benefiting at resolutions $256\times256$ and $512\times512$. Specifically for PixelGen and local editing, following the findings of Section \ref{subsec:profiles}, we consider an additional baseline, augmenting $\Delta\boldsymbol{v}_{\boldsymbol{\theta}}(\cdot,t,\boldsymbol{c})$ in Equation \ref{eq:earlyode} via a masking oracle for better localization as:
\begin{equation}
    [\Delta\boldsymbol{v}_{\boldsymbol{\theta},M,\tau}(\cdot,t,\boldsymbol{c})]_u:=
    \begin{cases}
        [\Delta\boldsymbol{v}_{\boldsymbol{\theta}}(\cdot,t,\boldsymbol{c})]_u & \text{if } t\geq\tau \text{ or } u\in\Omega_{M} \text{ and } t<\tau \\[8pt]
        0 & \text{otherwise}.
    \end{cases}
\end{equation}

\begin{table*}[t]
\centering
\caption{Editing with Qwen, FLUX.2 and PixelGen. Without CFG we tend to observe strong preservation but insufficient alignment. That is, CFG is often essential for image editing. However, recommended weights, $\omega$, tend to improve alignment while harming preservation. Early correction mitigates this tradeoff by largely improving preservation while maintaining strong alignment for Qwen-Image-Edit and PixelGen. For FLUX.2, found relatively conservative at $1024\times1024$ resolution, it recovers part of the distortion introduced by CFG while retaining its alignment gains. Moreover, our intervention is a key ingredient for unlocking robust masked guidance, i.e., removing it in the last row of the PixelGen experiments introduces severe distortion.
}
\label{tab:results}

\begingroup
\small
\setlength{\tabcolsep}{3pt}
\renewcommand{\arraystretch}{1.08}

\begin{tabular}{@{}>{\centering\arraybackslash}m{1.35em}@{\hspace{3pt}}p{5.5cm}*{6}{c}@{}}
\toprule
& & \multicolumn{3}{c}{Preservation}
& \multicolumn{3}{c}{Alignment} \\
\cmidrule(lr){3-5}
\cmidrule(l){6-8}
& Method
& {PSNR $\uparrow$}
& {LPIPS $\downarrow$}
& {DINOv2 $\uparrow$}
& {CLIP $\uparrow$}
& {PickScore $\uparrow$}
& {HPSv3 $\uparrow$} \\
\midrule

\multirow{6}{1.35em}{\centering\rotatebox[origin=c]{90}{\notsotiny $1024{\times}1024$}}
& Qwen-Image-Edit
& \textcolor{mygray}{13.06} & \textcolor{mygray}{0.4948} & \textcolor{mygray}{0.7082} & \textcolor{mygray}{0.2853} & \textcolor{mygray}{22.15} & \textcolor{mygray}{7.65} \\
& \quad \addrow Classifier-free guidance \hfill \textcolor{mygray}{($\omega=4.00$)}
& 11.25 & 0.5616 & 0.5750 & 0.2905 & 22.03 & 6.88 \\
& \quad \bluecell \addrow Early correction \hfill \textcolor{mygray}{($\tau=0.10$)}
& \bluecell \bfseries 14.05 & \bluecell \bfseries 0.4159 & \bluecell \bfseries 0.6606 & \bluecell \bfseries 0.2946 & \bluecell \bfseries 22.28 & \bluecell \bfseries 7.97 \\

\cmidrule{2-8}

& FLUX.2 [klein] 4B Base
& \textcolor{mygray}{27.65} & \textcolor{mygray}{0.0516} & \textcolor{mygray}{0.9926} & \textcolor{mygray}{0.2389} & \textcolor{mygray}{21.63} & \textcolor{mygray}{7.92} \\
& \quad \addrow Classifier-free guidance \hfill \textcolor{mygray}{($\omega=4.00$)}
& 20.22 & 0.1973 & 0.9163 & \bfseries 0.2605 & \bfseries 21.88 & 7.91 \\
& \quad \bluecell \addrow Early correction \hfill \textcolor{mygray}{($\tau=0.01$)}
& \bluecell \bfseries 20.33 & \bluecell \bfseries 0.1948 & \bluecell \bfseries 0.9171 & \bluecell 0.2603 & \bluecell \bfseries 21.88 & \bluecell \bfseries 7.92 \\

\midrule

\multirow{8}{1.35em}{\centering\rotatebox[origin=c]{90}{\notsotiny $512{\times}512$}}
& FLUX.2 [klein] 4B Base
& \textcolor{mygray}{14.07} & \textcolor{mygray}{0.4456} & \textcolor{mygray}{0.6496} & \textcolor{mygray}{0.2791} & \textcolor{mygray}{21.88} & \textcolor{mygray}{8.03} \\
& \quad \addrow Classifier-free guidance \hfill \textcolor{mygray}{($\omega=4.00$)}
& 12.05 & 0.4576 & 0.5973 & 0.2891 & 22.08 & 8.34 \\
& \quad \bluecell \addrow Early correction \hfill \textcolor{mygray}{($\tau=0.10$)}
& \bluecell \bfseries 13.81 & \bluecell \bfseries 0.3523 & \bluecell \bfseries 0.6828 & \bluecell \bfseries 0.2917 & \bluecell \bfseries 22.19 & \bluecell \bfseries 8.69 \\

\cmidrule{2-8}

& PixelGen-XXL
& \textcolor{mygray}{19.60} & \textcolor{mygray}{0.1625} & \textcolor{mygray}{0.8647} & \textcolor{mygray}{0.2798} & \textcolor{mygray}{22.62} & \textcolor{mygray}{7.15} \\
& \quad \addrow Classifier-free guidance \hfill \textcolor{mygray}{($\omega=4.00$)}
& 16.19 & 0.1818 & 0.8699 & \bfseries 0.2970 & 22.88 & 8.10 \\
&\quad \bluecell \addrow Early correction \hfill \textcolor{mygray}{($\tau=0.05$)}
& \bluecell 20.62 & \bluecell 0.1118 & \bluecell 0.9159 & \bluecell 0.2907 & \bluecell 22.94 & \bluecell 8.64 \\
& \quad \bluecell \addrow Masked guidance
& \bluecell \bfseries 21.54 & \bluecell \bfseries 0.0960 & \bluecell \bfseries 0.9186 & \bluecell 0.2888 & \bluecell \bfseries 22.98 & \bluecell \bfseries 8.73 \\
& \quad \textcolor{mygray}{\subrow Early correction}
& \textcolor{mygray}{13.93} & \textcolor{mygray}{0.3179} & \textcolor{mygray}{0.8430} & \textcolor{mygray}{0.2927} & \textcolor{mygray}{22.60} & \textcolor{mygray}{7.24} \\

\midrule

\multirow{3}{1.35em}{\centering\rotatebox[origin=c]{90}{\notsotiny $256{\times}256$}}
& FLUX.2 [klein] 4B Base
& \textcolor{mygray}{14.62} & \textcolor{mygray}{0.3668} & \textcolor{mygray}{0.6573} & \textcolor{mygray}{0.2638} & \textcolor{mygray}{21.42} & \textcolor{mygray}{5.18} \\
& \quad \addrow Classifier-free guidance \hfill \textcolor{mygray}{($\omega=4.00$)}
& 12.04 & 0.4461 & 0.5605 & 0.2858 & 21.91 & 7.30 \\
& \quad \bluecell \addrow Early correction \hfill \textcolor{mygray}{($\tau=0.10$)}
& \bluecell \bfseries 13.09 & \bluecell \bfseries 0.3850 & \bluecell \bfseries 0.6298 & \bluecell \bfseries 0.2884 & \bluecell \bfseries 22.04 & \bluecell \bfseries 7.84 \\

\bottomrule
\end{tabular}

\endgroup
\end{table*}

The masking strategy, however, is not sufficient. In fact, its benefits are only realized once our correction is applied and, counterintuitively, it otherwise leads to severe distortion in the outputs. This serves as evidence that early correction is a key ingredient in disentangling CFG dynamics. In particular, with reference to the left in Figure \ref{fig:tradeoff}, SAM 3 masks appear to offer minor gains beyond our proposed correction and tend to merely shift operating points without changing the underlying Pareto front.  We interpret this to mean that our correction alone leads to better localization, effectively approximating the masked oracle.

\begin{figure}
    \centering
    \includegraphics[width=\linewidth]{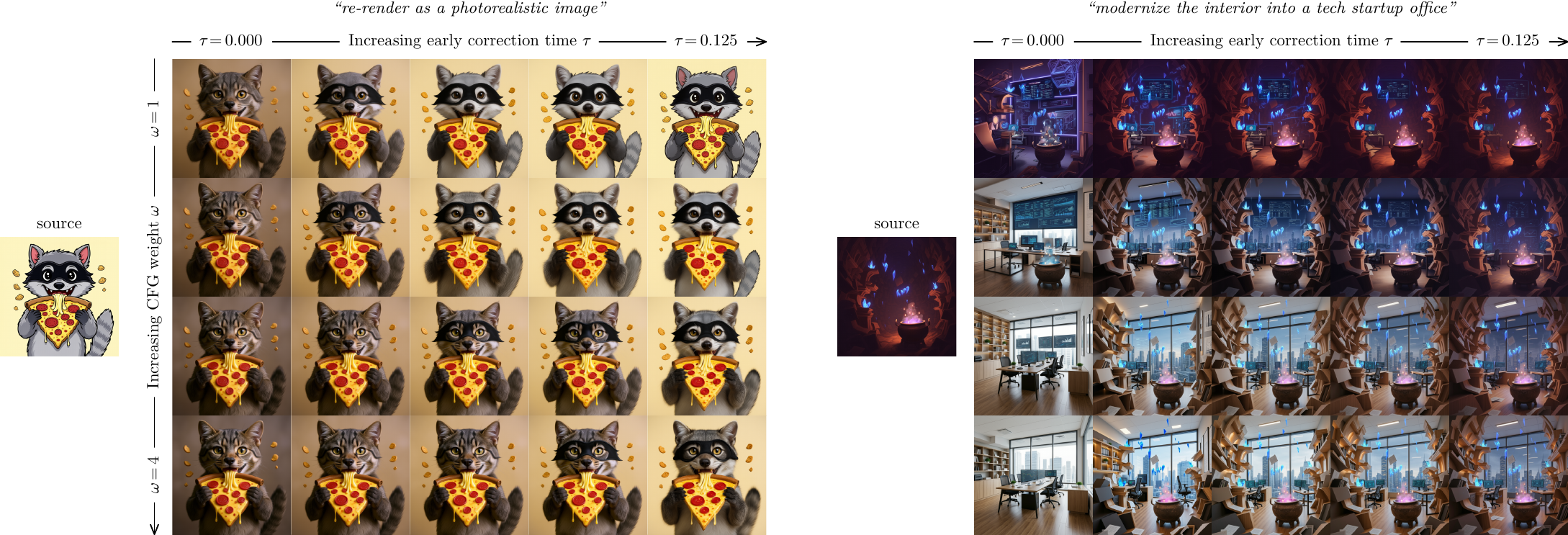}
    \caption{Qualitative analysis showing the effect of $\omega$ and $\tau$ at $1024
    \times1024$ resolution. On the left we show edits with Qwen-Image-Edit and on the right with FLUX.2 [klein] 4B Base. The first columns use $\tau=0$, i.e., no early correction. In both cases, $\tau=0$ tends to give generic results at large $\omega$ that may disregard the source. That is, on the left the raccoon is changed to a cat and on the right, eventually, no elements from the source survive. The proposed correction, activated when $\tau>0$, produces edits that are more faithful to the source without compromising instruction-following ability. See Figure \ref{fig:grid1} in Appendix \ref{appendix:moreexamples} for more examples.}
    \label{fig:grid}
\end{figure}

\begin{wrapfigure}{r}{0.6\textwidth}
    \centering
    \includegraphics[width=0.6\textwidth]{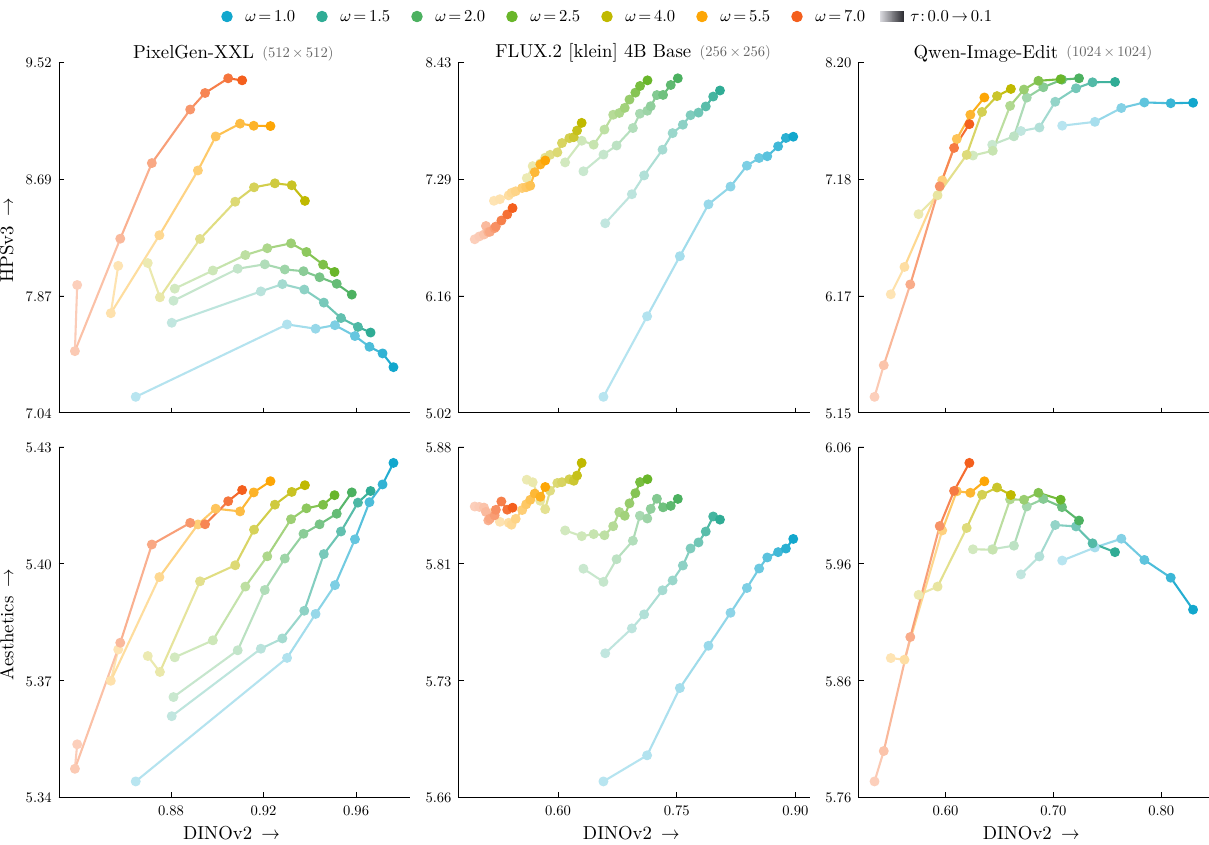}
    \caption{We show DINOv2-HPSv3, Aesthetics tradeoffs for the PixelGen, FLUX.2 and Qwen models with different $\omega$ and $\tau$. Curves represent fixed $\omega$ and form by varying $\tau$, with faint colors given to small values. A positive $\tau$ often improves all metrics.}
   \label{fig:tau}
\end{wrapfigure}

To show our approach is operationally distinct from CFG, we also explored the choice of early correction time $\tau$. Figure \ref{fig:grid} gives a qualitative summary of our findings, where small $\tau$ yield strong alignment but tend to also give somewhat generic results at large CFG weight $\omega$. Instead, large $\tau$ tend to maintain alignment while grounding edits to the source more strongly. Crucially, this does not necessarily compromise their integrity. Indeed, Figure \ref{fig:tau} shows our correction may simultaneously improve DINOv2, HPSv3 and Aesthetics scores.

\section{Discussion}

We studied CFG in TI2I flows and argued the recipe of \citet{ho2021classifierfree} is poorly integrated. Our observation is that CFG trains the text-unconditional branch to approximate the mean edit rather than a no-edit flow, so the mechanism responsible for alignment implicitly erodes preservation. We localized this effect to the noisy phase of modeling and proposed an intervention that substitutes a no-edit velocity into the unconditional branch for $t<\tau$. It requires no additional evaluations and recovers markedly richer alignment-preservation tradeoffs.

We do not present a rule for $\tau$. In practice, however, a uniform choice delivers consistent improvements. Moreover, we note that open TI2I models often do not provide complete information about their development, which may involve post-training we cannot account for. In these cases, we cannot cleanly attribute the success or relative ineffectiveness of our intervention to flow modeling alone.

Our findings caution against blind transfer of techniques across tasks. Despite its ubiquity, CFG is not a silver bullet and its naive application is harmful for source preservation in editing. We therefore view our work as an invitation to rethink such recipes. Instead of treating editing as T2I generation with an additional conditioning signal, we argue that TI2I modeling benefits from task-specific context, here a no-edit prior grounding generation to the source, and hope this perspective informs the design of future systems.

\bibliography{main}
\bibliographystyle{bst}

\newpage

\appendix

\begin{figure*}[t!]
    \centering

    \subfloat[Local editing with PixelGen]{{{\includegraphics[width=0.478\textwidth]{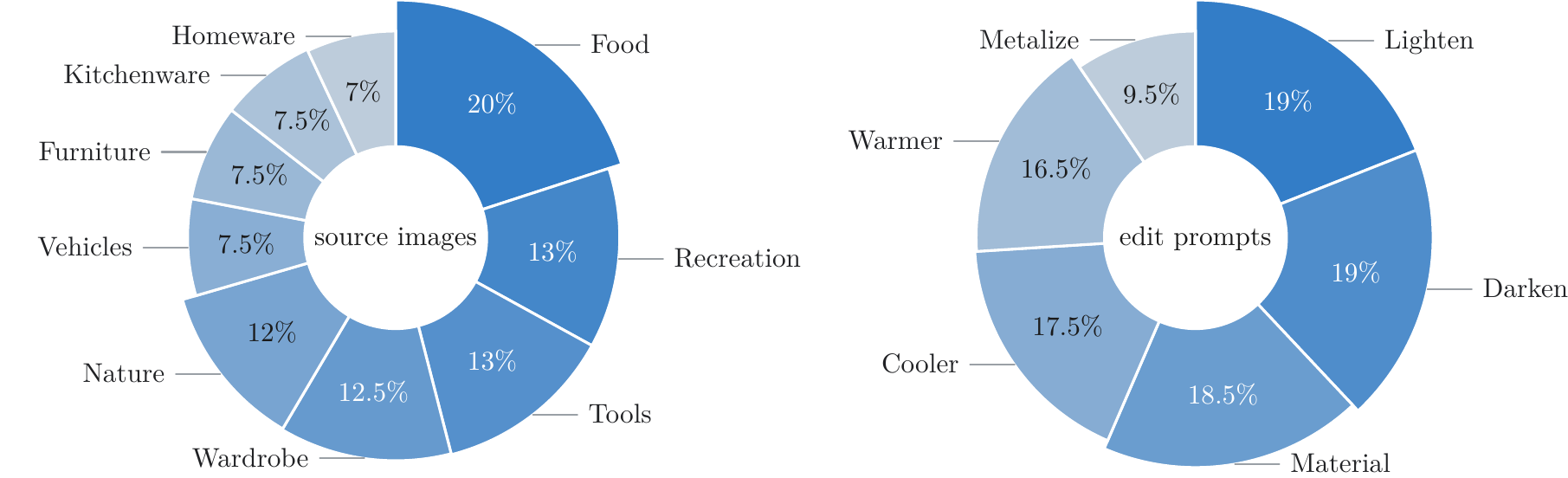}}}} \qquad
    \subfloat[General editing with Qwen and FLUX.2]{{{\includegraphics[width=0.478\textwidth]{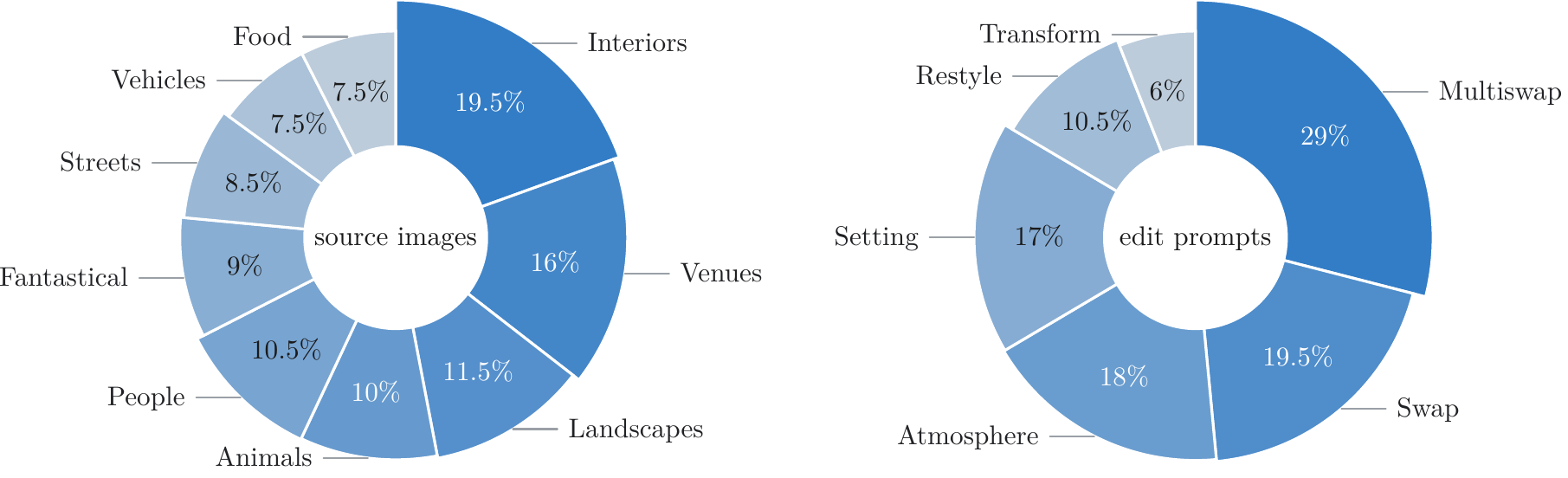}}}}
    \caption{Compositions of our two editing benchmarks, with approximately 200 entries each. The panels pair two donuts, showing the distribution of source images on the left and of edit prompts on the right. The benchmark on the left panel targets single objects on plain backdrops with localized recolor and material edits, i.e., confined within a specific region of the image. The benchmark on the right targets complex, typically crowded scenes and was specifically designed as a stress-test of TI2I capabilities. Together, they cover the spectrum from precise local edits to scene-wide changes that must still preserve surrounding context.}

   \label{fig:data}
\end{figure*}

\section{Experimental Setup}
\label{appendix:setup}

We ran inference on Linux machines using RTX Quadro 6000, RTX A5000, H100 GPUs and CUDA 13.0.

Benchmarks were generated via Claude Opus 4.7, FLUX.2 [klein] 4B Base \citep{flux-2-2025} ($1024\times1024$, 50 steps, $\omega=4$, Euler) and SAM 3 \citep{carion2026sam}, with compositions in Figure \ref{fig:data}.

For TI2I editing, which consumed most of our compute budget, we used the Euler solver with 25 steps. Preliminary runs with the recommended 50 steps showed little difference and did not justify the extra cost.

Regarding our pixel-space experiments, we finetuned PixelGen-XXL \citep{ma2026pixelgenimprovingpixeldiffusion} ($512\times512$, bf16) on NHR-Edit \citep{Kuprashevich_2026_WACV}, OmniEdit \citep{wei2025omniedit}, X2I \citep{omnigen} datasets for TI2I modeling and on Fine-T2I \citep{ma2026finet2iopenlargescalediverse} for T2I preservation. We used the JiT flow-matching velocity loss from \citet{Li_2026_CVPR} and LPIPS, DINOv2, REPA \citep{yu2025representation} regularizers applied similarly to the original PixelGen recipe, gated at $t=0.5$. We additionally applied unconditional T2I regularization and a KL penalty ($\beta=$1e-3) against a frozen reference to preserve base generative quality. The model was trained with the AdamW optimizer, a learning rate of 1e-4 and a batch size of 128 for 20k iterations.

We note that our adaptation of PixelGen reuses the T2I backbone without introducing any new parameters. In particular, source conditioning is realized through token concatenation, following the Kontext paradigm \citep{labs2025flux1kontextflowmatching}. That is, source images are embedded like the noisy inputs and their tokens are appended to the noisy token sequence inside the transformer. To distinguish them, noisy tokens carry the denoising timestep whereas source tokens are marked as noise-free, i.e., $t=1$. With this setup, information therefore propagates through the model's existing joint attention, requiring no adapters or auxiliary encoders.

\section{Additional Results}
\label{appendix:moreexamples}

\begin{figure}[!htb]
    \centering
    \includegraphics[width=\linewidth]{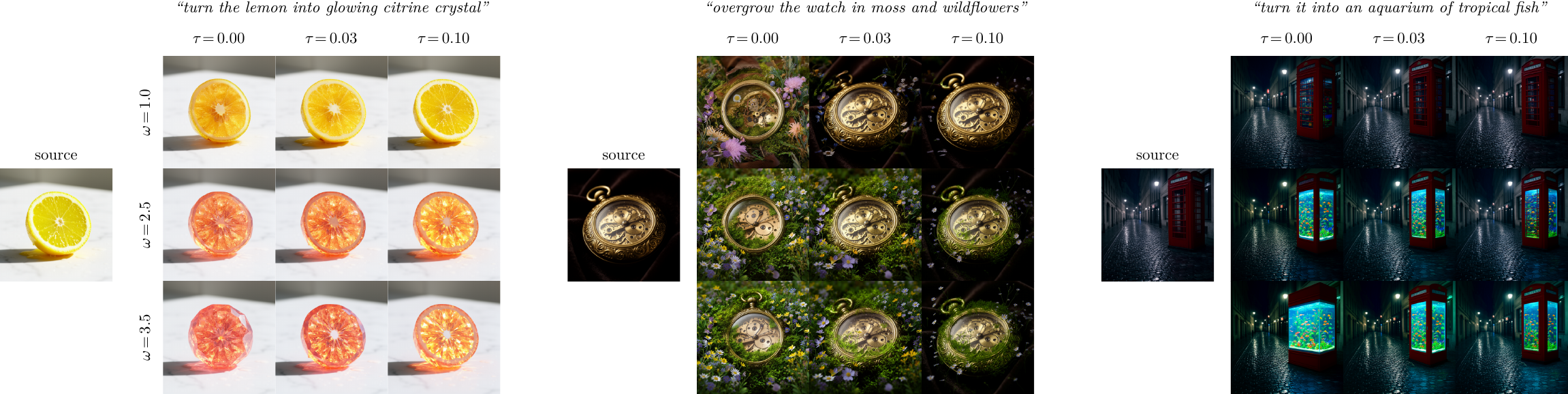}
    \caption{Editing with FLUX.2, showing the effect of $\omega$ and $\tau$ at $512
    \times512$ resolution. Ours, with $\tau>0$, maintains the lemon shape (left), perspective (middle) and booth shape (right) while adhering to the text.}
    \label{fig:grid1}
\end{figure}

\end{document}